\documentclass[letterpaper]{article} 
\usepackage{aaai25}  
\usepackage{times}  
\usepackage{helvet}  
\usepackage{courier}  
\usepackage[hyphens]{url}  
\usepackage{graphicx} 
\usepackage{natbib}  
\usepackage{caption} 
\usepackage{algorithm}
\usepackage{algorithmic}

\usepackage{newfloat}
\usepackage{listings}
\DeclareCaptionStyle{ruled}{labelfont=normalfont,labelsep=colon,strut=off} 
\floatstyle{ruled}
\newfloat{listing}{tb}{lst}{}
\floatname{listing}{Listing}
\title{CUPCase: Clinically Uncommon Patient Cases and Diagnoses Dataset}
\author{
    Oriel Perets\equalcontrib,
    Ofir Ben Shoham\equalcontrib,
    Nir Grinberg,
    Nadav Rappoport
}
\affiliations{
    Ben-Gurion University of the Negev\\

    Be'er Sheva, Israel\\
    {orielpe, benshoho}@post.bgu.ac.il; {nirgrn, nadavrap}@bgu.ac.il
}

\usepackage{bibentry}

\begin{document}

\maketitle

\begin{abstract}

Medical benchmark datasets significantly contribute to developing Large Language Models (LLMs) for medical knowledge extraction, diagnosis, summarization, and other uses.
Yet, current benchmarks are mainly derived from exam questions given to medical students or cases described in the medical literature, lacking the complexity of real-world patient cases that deviate from classic textbook abstractions. These include rare diseases, uncommon presentations of common diseases, and unexpected treatment responses. 
Here, we construct \textbf{C}linically \textbf{U}ncommon \textbf{P}atient \textbf{C}ases and Diagnosis Dataset (CUPCase) based on 3,562 real-world case reports from BMC, including diagnoses in open-ended textual format and as multiple-choice options with distractors.
Using this dataset, we evaluate the ability of state-of-the-art LLMs, including both general-purpose and Clinical LLMs, to identify and correctly diagnose a patient case, and test models' performance when only partial information about cases is available.
Our findings show that general-purpose GPT-4o attains the best performance in both the multiple-choice task (average accuracy of 87.9\%) and the open-ended task (BERTScore F1 of 0.764), outperforming several LLMs with a focus on the medical domain such as Meditron-70B and MedLM-Large. 
Moreover, GPT-4o was able to maintain 87\% and 88\% of its performance with only the first 20\% of tokens of the case presentation in multiple-choice and free text, respectively,  highlighting the potential of LLMs to aid in early diagnosis in real-world cases.
CUPCase expands our ability to evaluate LLMs for clinical decision support in an open and reproducible manner. 
\end{abstract}

%

\section{Introduction}
Large Language Models (LLMs) have demonstrated promising results in the medical field \cite{zhou2023survey}, including when general-purpose LLMs (such as GPT-4) are applied to medical tasks. For example, GPT-4 showed promising performance in the United States Medical Licensing Examination \cite{nori2023capabilities}, interpreting medical concepts (diagnoses, procedures, and drug codes) \cite{shoham2024medconceptsqa}, rare disease prediction \cite{do2024assessing}, and more.

Clinical LLMs (CLLMs) are specialized LLMs focused on the clinical domain, as opposed to general-purpose LLMs. Examples of CLLMs include Meditron \cite{chen2023meditron}, MedLM \cite{singhal2023large}, BioMistral \cite{labrak2024biomistral}, and Llama3-OpenBioLLM \cite{OpenBioLLMs}. CLLMs have several applications, such as disease prediction \cite{shoham2024cpllm}, medical chatbots, healthcare education, text generation, and more \cite{he2024foundation}.

To evaluate the performance of LLMs and CLLMs, medical clinical benchmarks are required. Notable examples of existing benchmarks in the medical domain include PubMedQA \cite{jin2019pubmedqa}, which focuses on question-answering over PubMed-derived questions; ClinicalBench \cite{liu2024large}, which covers tasks such as treatment recommendation and hospitalization summarization; and BioASQ-QA, which involves biomedical question-answering, document retrieval, text snippet extraction, and summarization \cite{krithara2023bioasq}. Additionally, some benchmarks are specifically designed for medical exam questions, such as MedQA \cite{jin2021disease} and MedMCQA, which are multiple-choice QA datasets tailored to medical entrance exam scenarios \cite{pal2022medmcqa}. Given their origin, these medical benchmarks often focus on evaluating LLMs' knowledge based on standard medical literature and fail to assess many skills necessary for deployment in a realistic clinical decision-making environment \cite{hager2024evaluation, mehandru2024evaluating}. In particular,  showed that only 5\% of reviewed works used real patient care data for LLM evaluation \cite{bedi2024systematic}. 

To complement these previous efforts, we propose using published case reports as a benchmark for LLMs. 
Clinical case reports are detailed accounts of individual patient cases that highlight unique or rare conditions, treatments, or outcomes, often providing insights into new or unusual aspects of medical practice. Published in scientific journals, they serve as valuable educational resources, contributing to medical knowledge by documenting specific clinical scenarios that may not be covered in larger studies.
Published patient case reports offer crucial insights beyond the scope of traditional textbooks. These reports can reveal significant scientific observations that might be missed in clinical trials, and expand our understanding by introducing novel findings and scenarios that deviate from classical cases \cite{cohen2006write}.
\begin{figure*}
    \centering
    \resizebox{\textwidth}{!}{ 
    \includegraphics{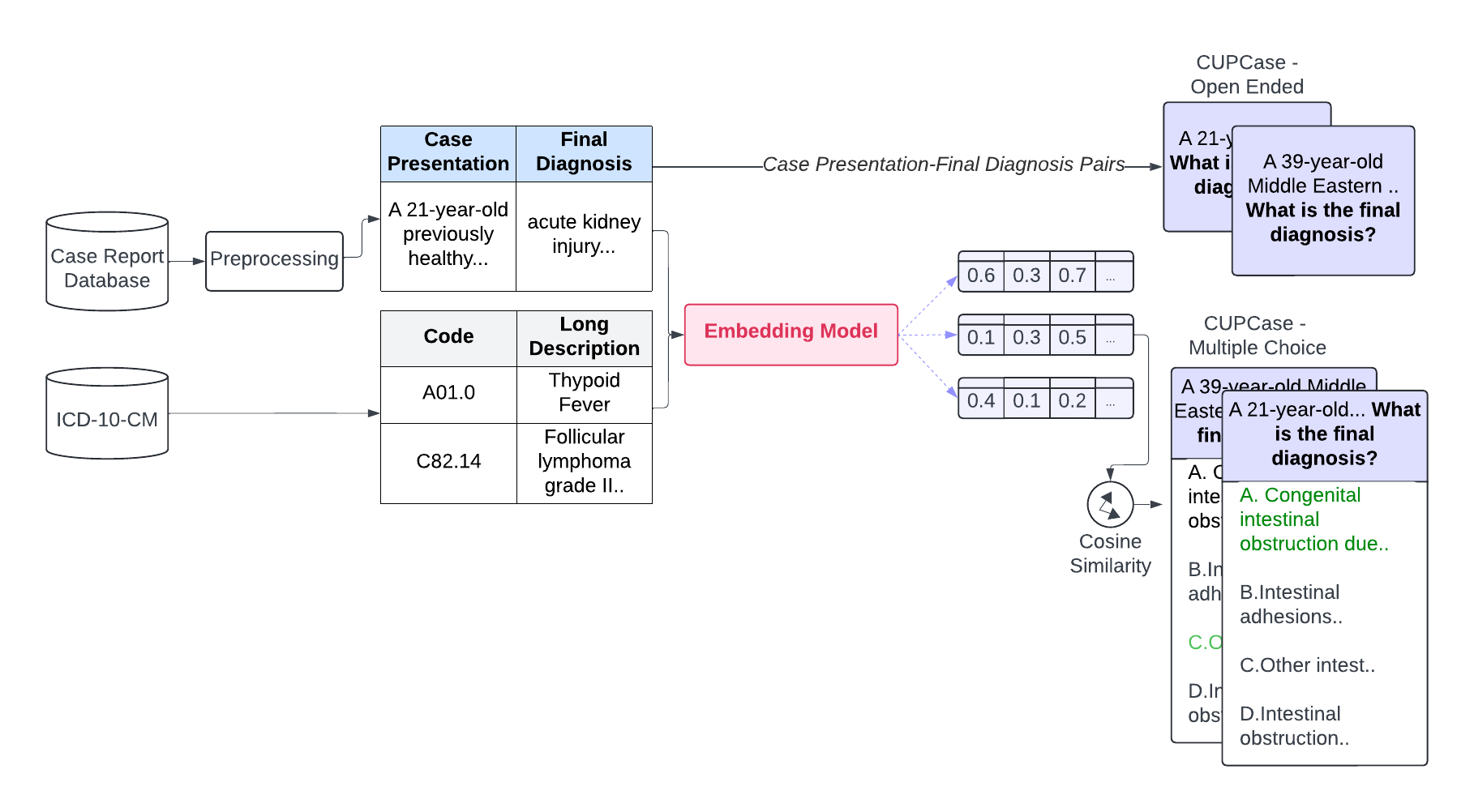}}
    \caption{Schematic flow of the dataset curation process.}
    \label{fig:method-flow}
\end{figure*}
\section{Related Work}
Several benchmarks were designed specifically to evaluate the performance of LLMs in rare diseases. For instance, \cite{reese2024evaluation} assessed the diagnostic accuracy of GPT-4 on 5,000 rare disease cases. RareBench \cite{chen2024rarebench} is a recent benchmark developed for rare diseases, demonstrating the potential of integrating LLMs into the clinical diagnostic process for these conditions. Another benchmark, RareDis \cite{martinez2022raredis}, includes 1,041 patient cases. Additionally, \cite{do2024assessing} found that GPT-4 and Claude achieved the highest performance in diagnosing rare diseases, based on evaluations of 200 synthetic patient cases and 275 publicly available patient cases. However, our proposed benchmark is not specific only to rare diseases.

Although existing previously published clinical benchmarks of rare diseases identification, these benchmarks do not address other complex patient cases, such cases may include uncommon presentations of common diseases or unexpected treatment responses. \cite{rios2024evaluation} examined complex medical cases from a single center (Massachusetts General Hospital Case Records) and asked GPT-4 to generate diagnoses. They found that GPT-4 provided the correct diagnosis on the first attempt in 42\% of the cases. However, they did not evaluate CLLMs, and their dataset consisted of only 75 medical cases. Another relevant work by \cite{kanjee2023accuracy} explored complex diagnostic identification beyond rare diseases. They applied GPT-4 to 70 complex cases from BMJ New England Journal of Medicine clinicopathologic conferences. Their dataset was extracted from these conferences and included cases with final pathological diagnoses used for educational purposes. They found that GPT-4 achieved an accuracy of 39\% on the first prediction attempt in their evaluation of open questions.

In contrast to previous studies, our work introduces a substantial dataset for evaluation, comprising 3,562 patient case presentations across a diverse array of clinical conditions, significantly expanding beyond the sample of about 75 cases in prior work. Moreover, our dataset is designed for scalability, allowing for easy expansion with additional case presentations from various journals using our generic open-source code. Importantly, our dataset encompasses a wide spectrum of scenarios, including rare diseases, uncommon presentations, and complex cases that deviate from classic medical knowledge. The dataset is also not confined to a single medical center, geographic location, or specific clinical practice (i.e. Rheumatology, Oncology, etc.).

Along with the newly constructed benchmark dataset, we conducted a thorough evaluation of current state-of-the-art LLMs, both specialized for the medical domain and general-purpose and included open and closed-source models. We evaluated the models in two tasks using zero-shot learning. The first task involved multiple-choice questions, where strong distractors were introduced by selecting choices with high semantic similarity. The second evaluation task required the models to generate a response to an open-ended diagnosis question, which emulates more closely the unconstrained nature of real-world diagnosis. In this approach, there are no options to choose from and the model is asked to generate the most probable diagnosis. In both evaluations, GPT-4o achieved the best performance: 87.9\% accuracy for multiple-choice questions and a BERTScore F1 of 0.7642 for free-text responses. Additionally, we found that general-purpose LLMs, such as GPT-4, outperformed CLLMs which were fine-tuned for the clinical domain.


\section{Method}
\subsection{Dataset Curation}
We extracted case reports from the BMC Journal of Medical Case Reports, spanning 2012 to 2020. We used a Python script to extract the ``case presentation'' section of each report, which includes free text and images.
To preprocess the dataset for evaluation, we utilized the GPT-4o-mini model through an API to remove any references to the final diagnosis and any follow-up treatments mentioned after the diagnosis, which we then validated.
We iteratively altered the prompt by preprocessing 35 random samples (~1\%) and manually evaluating the extraction and removal of the final diagnosis and the removal of any follow-up treatment. This step revealed limitations to the prompt, where the model would not follow instructions, or perform the removal of the final diagnoses from the text. Once the optimal prompt was selected, we applied the removal process to the entire dataset. The prompt used is:
\begin{quote}
\textit{
    Below is a case presentation of a patient, please remove any explicit reference to the final diagnosis from the text. 
    Additionally, remove any information about the patient's condition or treatment after the final diagnosis is made. 
    Do not remove any references to Figures or Images in the text like (Fig 1.)\\
    Return both the final diagnosis and the clean text separately as follows:\\
    Clean text: (clean text)\\
    Final diagnosis: (final diagnosis).\\
    Here is the Case presentation: \{case presentation\}
}

\end{quote}

We then performed another manual validation of the preprocessing step by randomly selecting 35 samples (~1\% dataset size) and validating the accurate extraction of the final diagnosis, removal of any explicit mention of the final diagnosis, and any follow-up treatment from the case presentation text.
This validation showed that all 35 randomly selected samples were successfully extracted for the final diagnosis and did not include any follow-up treatment. However, validation also revealed that 5 out of 35 (14\%) samples still explicitly mentioned the final diagnosis in the case presentation text.
To address this, we used another prompt explicitly mentioning the final diagnosis to be removed from the text:

\begin{quote}
\textit{
    ``Here is a case presentation, please remove any reference to the final diagnosis: \{final\_diagnosis\} 
    return the result as follows - Clean text: \textless clean\_text\textgreater\ end.\\\{case\_presentation\}''
}
\end{quote}
Further validation of another 35 randomly selected samples confirmed that all of these samples (~1\%) were successfully excluded for the final diagnosis, ensuring that the final diagnosis was no longer explicitly mentioned in the case presentation text.

\begin{figure}
    \centering
    \includegraphics[width=1\linewidth]{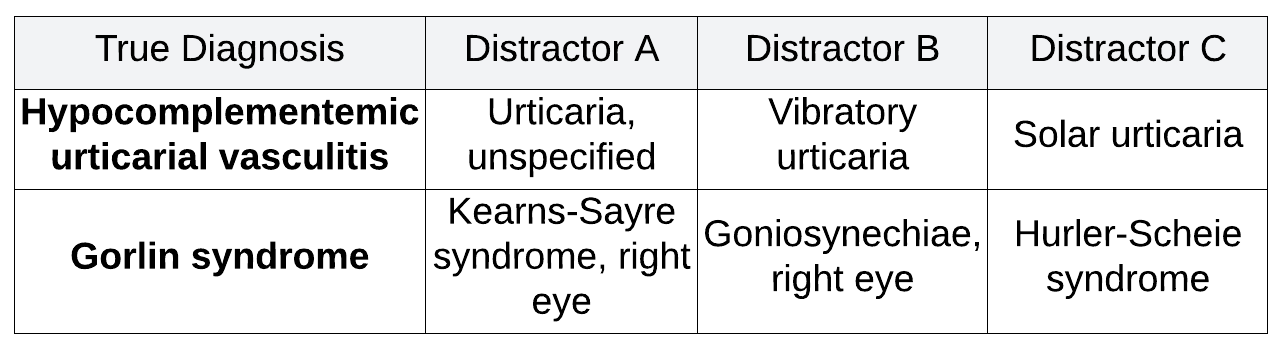}
        \captionof{table}{Sample of true diagnoses with ICD-10-CM codes' long description used as distractors.}
    \label{fig:icd-10-sample}
\end{figure}

Next, we employed the JINA AI embedding model \cite{gunther2023jina} to convert the final diagnosis into a 768-dimensional vector. This model was selected for its efficacy in capturing the semantics of short medical text embeddings, based on the findings of Excoffier et al. (2024) \cite{excoffier2024generalist}. Using the same model, we embedded the long descriptions of all 94,766 ICD-10-CM diagnoses into vectors of the same size. An example of the ICD-10-CM sample is presented in Table \ref{fig:icd-10-sample}.

The final dataset includes 3,562 patient cases, covering a wide variety of medical disciplines. The most prominent ones are Oncology (18.4\%), Infectious Diseases (10\%), and Neurology (9.88\%). Other substantial disciplines include Gastroenterology (5.17\%), Cardiology (4.27\%), and Obstetrics and Gynecology (4.39\%). A complete list is provided in Appendix A.

\begin{figure}[H]
    \centering
    \includegraphics[width=1\linewidth]{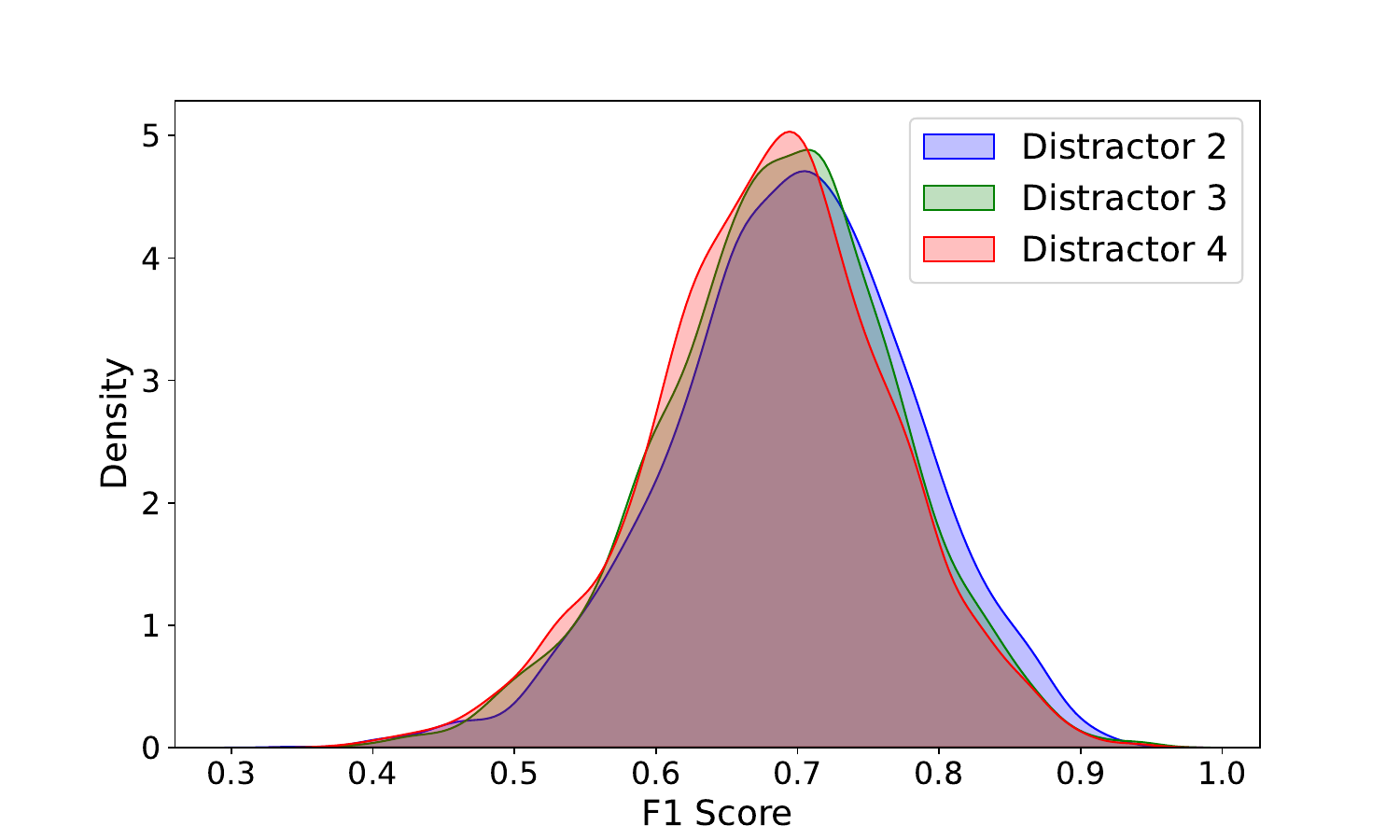}
    \caption{Distributions of BERTScore F1 between selected distractors and correct diagnoses, a BERTScore F1 of 1 indicates most similar, while 0 indicates least similar.}
    \label{fig:dist-bert-score}
\end{figure}

We calculated cosine similarity between the final diagnosis vector and the ICD-10-CM vectors to identify the top four most similar diagnoses to the final diagnosis. The second, third, and fourth closest diagnoses were set as distractors for the multiple-choice version of the dataset, under the assumption that the first match would likely be the correct diagnosis. This step was also validated in two ways: (a) randomly sampling 35 samples (~1\%) of the dataset and manually validating that the distractors do not contain the correct diagnosis, and (b) plotting the distribution of BERT scores F1 between the correct diagnoses and distractors selected, as shown in Figure \ref{fig:dist-bert-score}. The figure shows all three distractors are closely distributed with means 69.5\%, 68.6\%, 68\% for distractors 2, 3, and 4, respectively. This supports the claim that the distractors and correct diagnoses are semantically similar, as shown by the mean BERTScore F1 of the distrctors, rather than solely by the cosine similarity of their embedding vectors.

Ultimately, this process resulted in two datasets: one for QA, containing the correct final diagnosis and three distractors, and another with the free-text final diagnosis. The final dataset includes 3,562 patient cases and summary statistics about it are presented in Table \ref{tab:basic-statistics}. Although we focus on the text of cases, the dataset includes images as well, which facilitates the evaluation of multi-modal models in future work. 
The entire method flow is presented in Figure \ref{fig:method-flow}, visually representing our dataset curation and preprocessing steps.
\begin{table}[t]
\centering
\resizebox{0.95\columnwidth}{!}{
\begin{tabular}{lrr}
\hline
\textbf{Statistic} & \textbf{Case Presentation} & \textbf{Diagnosis}  \\
\hline
\# of samples & 3,562 & 3,271 \\
Minimum length &        69 & 2 \\
Maximum length &         3,876 & 86 \\
Average length &         938 & 11\\
Median length &         837 & 10\\
75th percentile &         1,160 & 14\\
95th percentile &         1,774  & 23\\\hline

\end{tabular}
}
\caption{General statistics for the dataset, \# of samples, indicates unique case presentations and unique diagnoses based on exact match only. Number of tokens for case presentation texts and final diagnosis texts in CUPCase, using Llama3-8B tokenizer, rounded to the nearest whole number.}
\label{tab:basic-statistics}
\end{table}

\section{Experiments}

We evaluated both general-purpose LLMs and CLLMs on our dataset, CUPCase. The general-purpose LLMs included in our evaluation are Llama3 and Llama3.1 (8B and 70B for both) \cite{dubey2024llama} and GPT-4o \cite{achiam2023gpt}. The CLLMs evaluated are BioMedGPT-LM-7B \cite{luo2023biomedgpt}, Meditron (7B and 70B) \cite{chen2023meditron}, Meerkat \cite{kim2024small}, BioMistral-7B-DARE \cite{labrak2024biomistral},  MedLM (Large) \cite{singhal2023large}, and Llama3-OpenBioLLM (8B and 70B), which continues instruction tuning on medical data from Llama3-70B-Instruct \cite{OpenBioLLMs}.
Our evaluation is based on zero-shot learning due to the length of the case reports. We conducted two experiments to evaluate various LLMs on CUPCase. The first experiment involved creating multiple-choice questions where the LLM chooses one option out of four. In the second experiment, we prompted the LLM to generate a diagnosis and then we evaluated the performance based on text similarity between the target disease and the predicted diagnosis.

\subsection{Multiple-Choice Evaluation}

In the first experiment, we evaluated CUPCase using a multiple-choice format. Each question consists of a general prompt, a patient case presentation, and four options: three distractors and one correct answer in a random order. Nearly all models (except GPT-4o and MedPalm) were evaluated using the lm-evaluation-harness \cite{eval-harness} python package on NVIDIA RTX6000 GPU. For these models, we chose the diagnosis with the highest probability out of the four options. The probability for each diagnosis is based on the likelihood of generating that diagnosis given the context of the text. For GPT-4o and MedPalm, we used API access to prompt the model to indicate the correct option (A, B, C, or D) from the provided choices. 

Our metric for the multiple-choice evaluation is accuracy, as the options are balanced by shuffling four options. GPT-4o achieved the highest accuracy with a score of 87.9\%. Following closely is Llama3-70B Instruct with an accuracy of 85.7\%. Other LLMs with similar performance include Meditron-70B at 85.55\% and Llama3.1-70B Instruct at 84.8\%. GPT-4o is the best general-purpose LLM, while Meditron-70B is the leading clinical LLM. Despite being designed as expert clinical models, all the CLLMs underperformed compared to some general-purpose models, including GPT-4o and Llama3-70B Instruct. Additionally, Llama3 OpenBioLLM 70B, which did continuous pre-training from Llama3-70B Instruct on medical instructions, achieved lower accuracy compared to Llama3-70B Instruct. Among the 7B/8B LLMs, Meditron-7B achieved the highest accuracy with 82.27\%, though this is still significantly lower than the 87.9\% achieved by GPT-4o. Results are presented in Table \ref{tab:all-results}.

\begin{table*}[t]
\centering
\small
\begin{tabular}{lrrrr}
\hline
\textbf{Model} & \textbf{Accuracy} & \textbf{Std} & \textbf{BERTScore F1} & \textbf{Std} \\ 
\hline
GPT-4o & \textbf{87.90} & 2.22 & \textbf{76.42} & 0.96 \\
Llama3-70B Instruct & 85.70 & 1.21 & 72.67 & 0.43 \\
Llama3.1-70B Instruct & 84.80 & 0.86 & 72.31 & 0.51 \\
Llama3.1-8B Instruct & 81.02 & 1.40 & 66.97 & 0.60 \\
Llama3-8B Instruct & 80.05 & 1.92 & 50.69 & 0.63 \\
\hline
\\[\medskipamount]
\multicolumn{5}{l}{\textbf{Clinical LLMs}} \\
\hline
Meditron-70B  & \textbf{85.55} & 0.92 & 59.77 & 0.30 \\
MedLM-Large  & 84.40 & 1.57 & \textbf{71.32} & 0.56 \\
Llama3 OpenBioLLM 70B  & 84.27 & 1.31 & 65.22 & 0.38 \\
Meditron-7B  & 82.27 & 1.37 & 59.19 & 0.59 \\
Bio Mistral  & 81.55 & 1.82 & 61.13 & 1.12 \\
BioMedGPT  & 79.62 & 1.08 & 65.22 & 0.38 \\
Llama3 OpenBioLLM 8B  & 77.75 & 1.79 & 64.29 & 0.78 \\
Meerkat  & 77.47 & 1.65 & 51.08 & 0.70 \\
\hline
\end{tabular}
\caption{Zero-shot evaluation results for multiple-choice question-answering. Mean accuracy and standard deviation were calculated using 8 bootstrap samples of 500 samples each. For GPT-4o and MedLM-Large, we utilized four bootstrap samples, each comprising 250 samples, due to cost constraints.}
\label{tab:all-results}
\end{table*}

\subsection{Free text Evaluation}
In the second experiment, we evaluate CUPCase using an approach that we believe is more suitable for real-world scenarios. As in real-world situations, there are no closed questions, but open-ended questions make more sense. Therefore, in the free-text evaluation, using zero-shot learning, we provide the LLMs with a case presentation and ask them to generate the predicted disease for the case presentation. For this task, the same CUPCase dataset was used. The results of this free text evaluation are presented in Table \ref{tab:all-results}. The metric is BERTScore F1, as we want to estimate based on semantic similarity \cite{zhang2019bertscore}. GPT-4o demonstrates the best results, with a BERTScore F1 of 76.42\%, showing a significant improvement compared to CLLMs such as MedLM-Large, which shows the highest results for this evaluation. Although Meditron-70B achieves the highest accuracy in multiple-choice evaluations on our dataset, it achieved a BERTScore F1 of only 59.77\%, which is lower than the 71.32\% of MedLM-Large and other CLLMs. Similarly, as observed previously, Llama3-OpenBioLLM-70B underperformed compared to Llama3-70B-Instruct, despite being a continuation of Llama3-70B-Instruct and being trained on medical instructions. BioMistral achieved a BERTScore F1 of 61.13\% and outperformed Meditron-70B in this experiment, despite having only 7 billion parameters compared to Meditron-70B's 70 billion parameters. Another interesting result is that Llama3.1-8B-Instruct achieved a BERTScore F1 of 66.97\%, compared to 50.69\% for Llama3-8B-Instruct.

\subsection{Cumulative Information Analysis}
To further investigate the diagnostic inference capabilities of general-purpose LLMs when processing medical case reports, we conducted a sensitivity analysis using a cumulative information approach. This involves incrementally providing the model with increasing percentages of tokens from the case presentation text (i.e. the first 20\% of the tokens in the case presentation, then the first 40\% and so on), allowing us to assess the relation between the amount of clinical information and the diagnosis accuracy of the model. This approach is justified by the fact a case presentation is written in chronological manner, describing the patient's timeline in the clinic throughout the diagnosis process \cite{cohen2006write}. 
The cumulative information analysis serves as an evaluation metric for LLMs' performance in patient diagnosis assistance. By simulating the iterative nature of the diagnostic process, where patient information is gathered progressively through various clinical means. This approach offers insights into the models' ability to formulate correct diagnoses with limited information.
Future work can use this analysis to explore the superiority or inferiority of the model compared to the physician. 

\begin{figure*}
    \centering
    \begin{minipage}{0.45\textwidth}
        \centering
        \includegraphics[width=\linewidth]{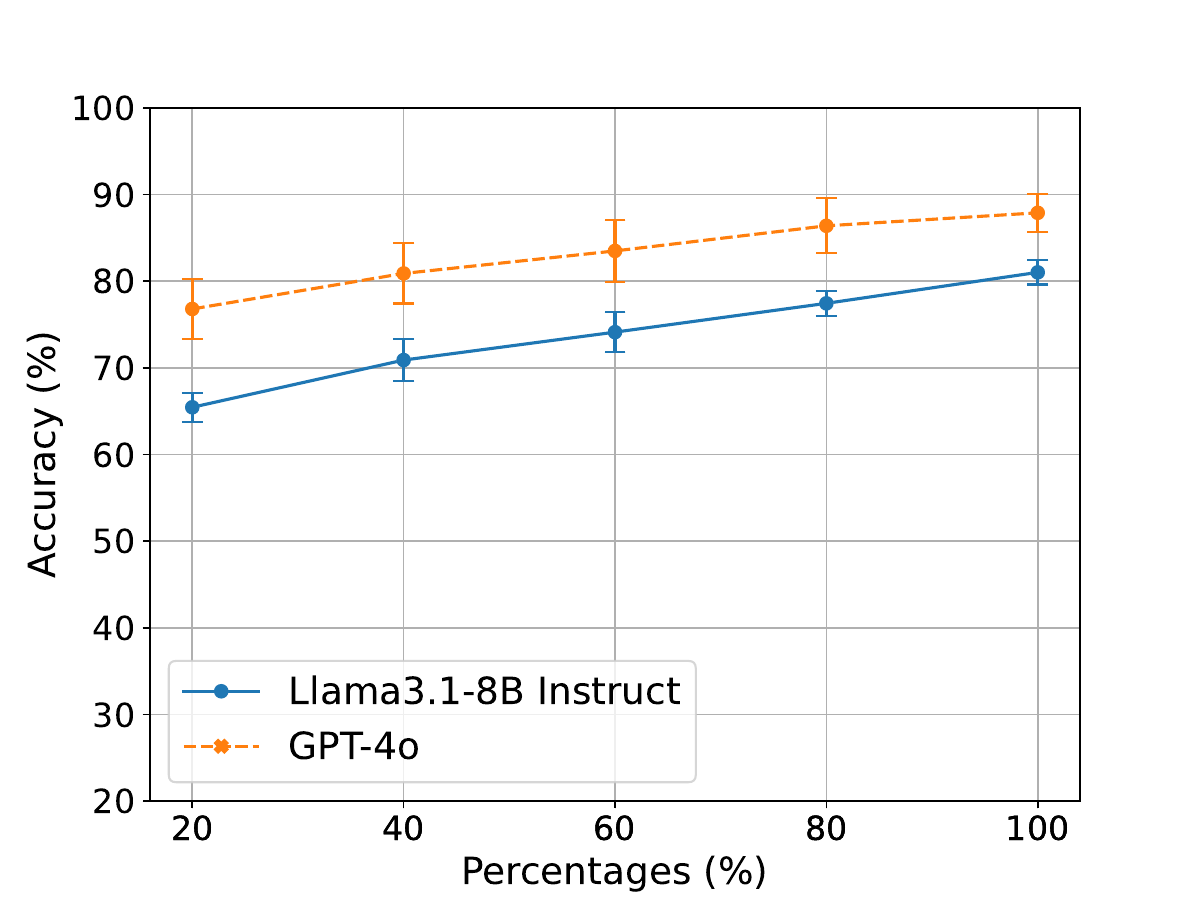}
        \caption{Accuracy per percentage of tokens in the case presentation text. The error bar represents the standard deviation.}
        \label{fig:ablation_accuracy_per_percentages}
    \end{minipage}\hfill
    \begin{minipage}{0.45\textwidth}
        \centering
        \includegraphics[width=\linewidth]{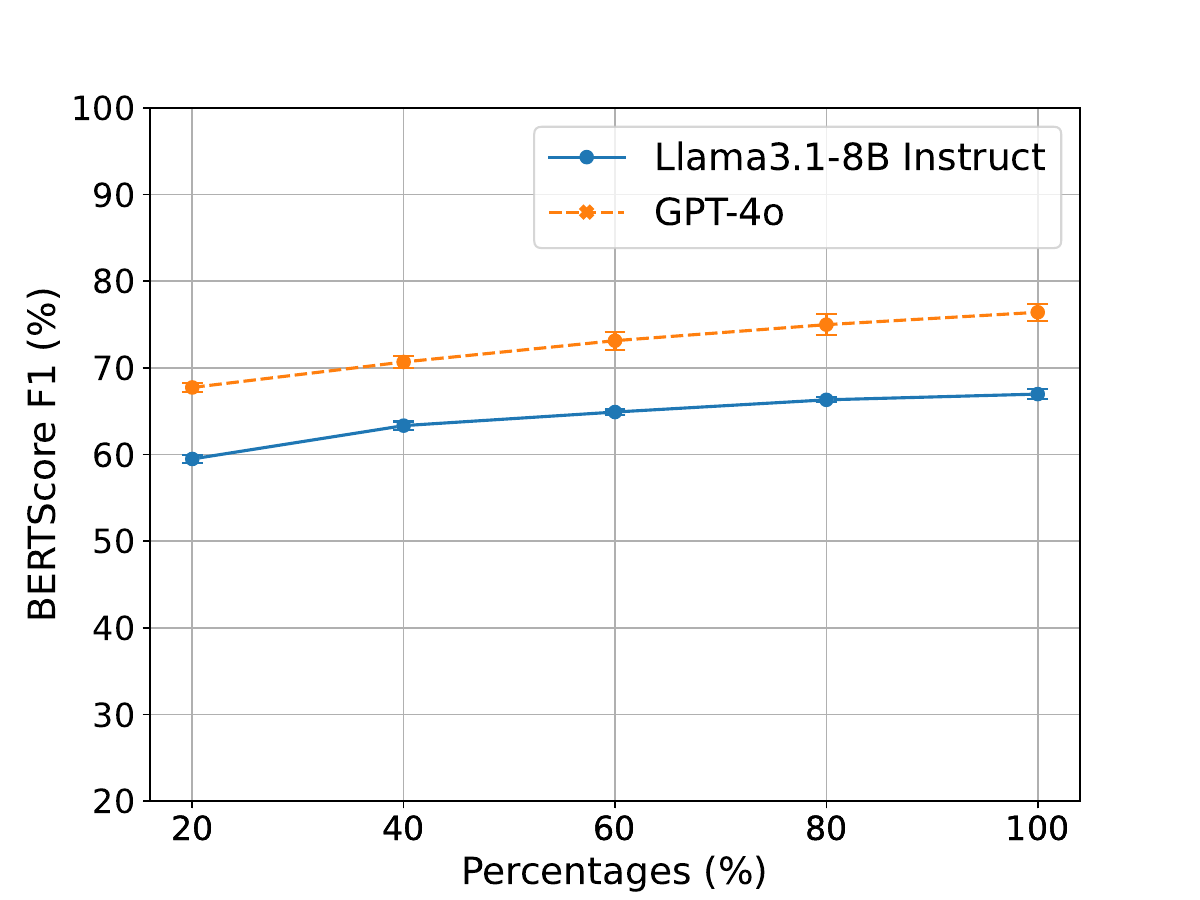}
        \caption{BERTScore F1 per percentage of tokens in the case presentation text. The error bar represents the standard deviation.}
        \label{fig:ablation_bert_score_per_percentages}
    \end{minipage}
\end{figure*}

Figures \ref{fig:ablation_accuracy_per_percentages} and \ref{fig:ablation_bert_score_per_percentages} show the performance of our two evaluations (multiple-choice and free-text) with varying percentages of tokens in the case presentation text, based on the Llama3.1-8B tokenizer. We show results for both the best-performing LLM (GPT-4o) and a smaller LLM (Llama3.1-8B Instruct). As illustrated in these graphs, and as expected, the performance improves as the percentage of the case presentation text considered increases, for both multiple-choice and free-text evaluations, indicating more context results in better diagnosis prediction. 
For the free text generation, GPT-4o was able to achieve an average BERT Score F1 of \(67.74\% \pm .5\), retaining 88.6\% of the performance using the entire case presentation text, with only 20\% of the tokens. Multiple-choice settings results (mean accuracy and standard deviation of \(76.8\% \pm 3.46\)) show similar results with GPT-4o retraining 87.3\% of performance with only 20\% of the tokens in the case presentation.
This reservation of performance is a good sign as it stimulates real-world scenarios, where context about the patient is incrementally added in the diagnosis process, supporting claims these models can provide diagnostic assistance.

\subsection{Error Analysis}
We performed an error analysis by providing two examples to two physicians to investigate and provide a deeper understanding of the model's behavior. For the error analysis, we used the best-performing LLM (GPT-4o). 

We employ two expert radiologists for the task, hence selecting two case reports from the dataset in the field of Rheumatology. We provide the physicians with a clean case presentation, the correct final diagnosis, and the model-generated diagnosis. We ask the physicians to describe why, in their medical opinion, the model mistakenly diagnosed the case, considering the reported patient background, symptoms, tests, procedures, and described imaging results. We also asked each physician to determine whether they believed a trained physician was likely to make the same mistake as the model. 

\textbf{Case 1:} The first case describes a 48-year-old Thai woman \cite{martviset2020urinary}, who was ultimately diagnosed with \textit{Balantidiasis} by the model, where the correct diagnosis was \textit{Clinically suspected SLE with lupus nephritis}. Considering this case, the first physician, an experienced Rheumatologist explained he expected the model to diagnose a lupus-related diagnosis correctly until Balantidium was repeatedly (three times) found in the patient's urine. The absence of further systematic lupus symptoms left the diagnosis unclear. Ultimately, the physician claims a diagnosis of Balantidiasis could explain many of the patient's symptoms and test results, including the anemia, the urine findings, and the patient's origin.

\textbf{Case 2:} The second case describes an 88-year-old Caucasian woman, who was ultimately diagnosed by the model with \textit{Wegener's Granulomatosis (now known as GPA)}, where the correct diagnosis was \textit{Polyarteritis Nodosa (PAN)} \cite{ortu2013polyarteritis}. Considering this case, the second physician, also an experienced Rheumatologist mentioned the diagnosis in this case is difficult even for an experienced physician. In his opinion, the model might have been mistaken due to several factors. First, ear, nose, and throat (ENT) involvement exists, which is much more common in GPA than PAN. Moreover, some key features of PAN were missed, specifically, necrosis of the palate and fibrinoid necrosis in a biopsy. Another factor is elevated blood pressure, which is prevalent in PAN but could be caused by other conditions. Kidney involvement is prevalent in both diagnoses could also contribute to the mistake. Last, in recent years, PAN was mostly associated with Hepatitis B (HBV) and sometimes Hepatitis C (HCV), but nowadays, HBV is mostly vaccinated. Hence, the model might have mistaken the negative HBV test as a negative predictor for PAN, steering the diagnosis in another direction.

Overall, this error analysis cases, while specific and addressing Rheumatology, can indicate the difficulty and intricacies of the constructed dataset. Furthermore, we believe the analysis shows a physician could have made the same mistaken diagnosis, and we, therefore, hypothesize the model's error may not come from a lack of knowledge but rather an abundance of other general knowledge in medicine. This preliminary conclusion can suggest the need for specialized models dealing with these uncommon cases as shown in previous studies \cite{liu2023tailoring}.


\section{Discussion}

Although being a general LLM, GPT-4o achieved the best performance in both multiple-choice and free-text evaluations. However, since GPT-4o is not open-source, its use in real clinical settings for diagnosing patient case presentations may pose privacy concerns.
For the multiple-choice evaluation, Llama3-70B Instruct, an open-source model, achieved the second-best performance. Unlike GPT-4o, Llama3-70B Instruct can be used locally without privacy issues. However, its mean accuracy is 2.2\% lower (absolute) than that of GPT-4o. The relative difference in accuracy is 2.57\%. All models demonstrated significantly higher accuracy compared to random guessing (25\%), indicating their capability to understand the task and provide meaningful answers in most cases.
In the free-text evaluation, Llama3-70B Instruct was the second-best model, with a mean BERTScore F1 score 3.75\% lower (absolute) than GPT-4o. The relative difference in accuracy is 5.16\%. Overall, GPT-4o was the top-performing model in both evaluations, with Llama3-70B Instruct coming in second.

Our evaluations on our proposed CUPCase dataset showed that general-purpose LLMs, such as GPT-4o and Llama3-70B Instruct, outperformed other models, including CLLMs that were specifically fine-tuned for the clinical domain using continuous pre-training or instruction-tuning. These results suggest that current state-of-the-art CLLMs may need further improvement compared to general-purpose LLMs, even for clinical tasks. This finding is consistent with other works, such as MedConceptsQA \cite{shoham2024medconceptsqa}, which presented similar claims for different clinical tasks.

LLMs with a high number of parameters can be challenging to deploy in production because of potentially high resources demands and relatively high costs. Moreover, particularly non-open-source models such as GPT-4o and MedLM-Large may not be used due to clinical data privacy concerns. As shown in the multiple-choice experiment, the highest accuracy among 'small' (7B-8B) LLMs is achieved by Meditron-7B, with an accuracy of 82.27\%. However, this result is 5.63\% lower in absolute accuracy compared to the best model, GPT-4o. In the free-text evaluation, Llama3.1-8B Instruct achieved a BERTScore F1 of 66.97\%, which is 9.45\% lower in absolute BERTScore F1 compared to GPT-4o. These results suggest that the smaller models offer lower performance, but these models provide alternatives in terms of privacy and reduced production costs.

In this paper, we proposed the CUPCase dataset primarily as a resource for evaluating language models' knowledge and diagnostic ability. Our promising evaluation results suggest that LLMs have the potential to be utilized to diagnose patient case presentations. For instance, doctors could use LLMs to assist with complex patient cases by obtaining possible diagnoses through free-text generation. Another potential application is using LLMs to identify target diseases for billing purposes and for alerting for misdiagnoses.

\subsection{Limitations}
Our work has several limitations. Our dataset contains case presentations solely from the BMC Journal of Medical Case Reports, and may not fully represent the entire clinical domain. However, our open-source code can be easily used to expand the dataset to other sources. Additionally, while we employ manual validation of randomly sampled examples of the CUPCase dataset, manually validating the entire dataset can further ensure the validity of all the samples. Moreover, we used only zero-shot learning for evaluation, as the texts are longer and contain an average of 938 tokens according to Llama3-8B tokenizer. However, this represents the real world where LLMs are limited by a number of input tokens. Additionally, we used only the text from the case presentations and did not include complementary information like figures. Another limitation is that we cannot be certain that the dataset case presentations were not included in the training data of the evaluated LLMs.

\subsection{Future Work}
This work sets the path for multiple future works. Given the multimodality of the dataset (which contains both texts and images), a natural future direction is to evaluate the performance of multi-modal language models on the multi-modal dataset. Future work, building upon this dataset can also conduct experiments concerning rare diseases specifically, which could provide insights into how well LLMs and Clinical LLMs perform for the diagnosis of rare cases. Additionally, As our error analysis indicates, challenging examples often require domain-specific expertise for accurate responses. Consequently, we think that a mixture-of-experts approach, where each expert is a fine-tuned model specializing in a specific clinical domain, may improve the diagnostic performance of case presentations \cite{cai2024survey}.

\subsection{Data and Code availability}
The complete dataset constructed in this study is available at the following link: 
\url{https://huggingface.co/datasets/ofir408/CupCase}. All code for preprocessing, dataset construction, and evaluations can be found in the corresponding GitHub repository: 
\url{https://github.com/nadavlab/CUPCase}.

\appendix
\section{Appendix A}

\subsection{Medical Disciplines in CUPCase}

\begin{table}[H]
\centering
\begin{tabular}{lrr}
\hline
\textbf{Medical Discipline} & \textbf{\% of Samples} \\ \hline
Oncology                    & 18.43                \\ \hline
Infectious Disease          & 10.00                 \\ \hline
Neurology                   & 9.88                  \\ \hline
Gastroenterology            & 5.17                  \\ \hline
Cardiology                  & 4.27                  \\ \hline
Obstetrics and Gynecology   & 4.39                  \\ \hline
Hematology                  & 4.16                  \\ \hline
Endocrinology               & 4.04                  \\ \hline
Orthopedics                 & 3.71                  \\ \hline
Ophthalmology               & 3.37                  \\ \hline
Rheumatology                & 2.81                  \\ \hline
Pulmonology                 & 2.47                  \\ \hline
Dermatology                 & 2.47                  \\ \hline
Emergency Medicine          & 2.13                  \\ \hline
Surgery                     & 2.13                  \\ \hline
Nephrology                  & 2.36                  \\ \hline
Psychiatry                  & 1.69                  \\ \hline
Pediatrics                  & 1.35                  \\ \hline
Allergy and Immunology      & 1.35                  \\ \hline
Urology                     & 1.24                  \\ \hline
Otolaryngology (ENT)        & 0.79                  \\ \hline
Internal Medicine           & 0.56                  \\ \hline
Neurosurgery                & 0.45                  \\ \hline
Anesthesiology              & 0.22                  \\ \hline
Physical Medicine and Rehabilitation & 0.22         \\ \hline
Vascular Surgery            & 0.11                  \\ \hline
Pathology                   & 0.11                  \\ \hline
Radiology                   & 0.11                  \\ \hline
\end{tabular}
\caption{Percentage of Samples by Medical Discipline (Sorted by Percentage)}
\label{table:medical_disciplines_sorted}
\end{table}

\section{Acknowledgments}
We gratefully acknowledge Mohammad Naffa, MD, and Amir Bieber, MD for their valuable assistance with the error analysis part of this study.
\bibliography{aaai25}

\begin{thebibliography}{36}
\providecommand{\natexlab}[1]{#1}

\bibitem[{Achiam et~al.(2023)Achiam, Adler, Agarwal, Ahmad, Akkaya, Aleman, Almeida, Altenschmidt, Altman, Anadkat et~al.}]{achiam2023gpt}
Achiam, J.; Adler, S.; Agarwal, S.; Ahmad, L.; Akkaya, I.; Aleman, F.~L.; Almeida, D.; Altenschmidt, J.; Altman, S.; Anadkat, S.; et~al. 2023.
\newblock Gpt-4 technical report.
\newblock \emph{arXiv preprint arXiv:2303.08774}.

\bibitem[{Ankit~Pal(2024)}]{OpenBioLLMs}
Ankit~Pal, M.~S. 2024.
\newblock OpenBioLLMs: Advancing Open-Source Large Language Models for Healthcare and Life Sciences.
\newblock \url{https://huggingface.co/aaditya/OpenBioLLM-Llama3-70B}.

\bibitem[{Bedi et~al.(2024)Bedi, Liu, Orr-Ewing, Dash, Koyejo, Callahan, Fries, Wornow, Swaminathan, Lehmann et~al.}]{bedi2024systematic}
Bedi, S.; Liu, Y.; Orr-Ewing, L.; Dash, D.; Koyejo, S.; Callahan, A.; Fries, J.~A.; Wornow, M.; Swaminathan, A.; Lehmann, L.~S.; et~al. 2024.
\newblock A Systematic Review of Testing and Evaluation of Healthcare Applications of Large Language Models (LLMs).
\newblock \emph{medRxiv}, 2024--04.

\bibitem[{Cai et~al.(2024)Cai, Jiang, Wang, Tang, Kim, and Huang}]{cai2024survey}
Cai, W.; Jiang, J.; Wang, F.; Tang, J.; Kim, S.; and Huang, J. 2024.
\newblock A survey on mixture of experts.
\newblock \emph{arXiv preprint arXiv:2407.06204}.

\bibitem[{Chen et~al.(2024)Chen, Mao, Guo, Wang, Zhang, and Chen}]{chen2024rarebench}
Chen, X.; Mao, X.; Guo, Q.; Wang, L.; Zhang, S.; and Chen, T. 2024.
\newblock RareBench: Can LLMs Serve as Rare Diseases Specialists?
\newblock \emph{arXiv preprint arXiv:2402.06341}.

\bibitem[{Chen et~al.(2023)Chen, Cano, Romanou, Bonnet, Matoba, Salvi, Pagliardini, Fan, K{\"o}pf, Mohtashami et~al.}]{chen2023meditron}
Chen, Z.; Cano, A.~H.; Romanou, A.; Bonnet, A.; Matoba, K.; Salvi, F.; Pagliardini, M.; Fan, S.; K{\"o}pf, A.; Mohtashami, A.; et~al. 2023.
\newblock Meditron-70b: Scaling medical pretraining for large language models.
\newblock \emph{arXiv preprint arXiv:2311.16079}.

\bibitem[{Cohen(2006)}]{cohen2006write}
Cohen, H. 2006.
\newblock How to write a patient case report.
\newblock \emph{American Journal of Health-System Pharmacy}, 63(19): 1888--1892.

\bibitem[{do~Olmo et~al.(2024)do~Olmo, Logrono, Mascias, Martinez, and Isla}]{do2024assessing}
do~Olmo, J.; Logrono, J.; Mascias, C.; Martinez, M.; and Isla, J. 2024.
\newblock Assessing DxGPT: Diagnosing Rare Diseases with Various Large Language Models.
\newblock \emph{medRxiv}, 2024--05.

\bibitem[{Dubey et~al.(2024)Dubey, Jauhri, Pandey, Kadian, Al-Dahle, Letman, Mathur, Schelten, Yang, Fan et~al.}]{dubey2024llama}
Dubey, A.; Jauhri, A.; Pandey, A.; Kadian, A.; Al-Dahle, A.; Letman, A.; Mathur, A.; Schelten, A.; Yang, A.; Fan, A.; et~al. 2024.
\newblock The Llama 3 Herd of Models.
\newblock \emph{arXiv preprint arXiv:2407.21783}.

\bibitem[{Excoffier et~al.(2024)Excoffier, Roehr, Figueroa, Papaaioannou, Bressem, and Ortala}]{excoffier2024generalist}
Excoffier, J.-B.; Roehr, T.; Figueroa, A.; Papaaioannou, M.; Bressem, K.; and Ortala, M. 2024.
\newblock Generalist embedding models are better at short-context clinical semantic search than specialized embedding models.
\newblock \emph{arXiv preprint arXiv:2401.01943}.

\bibitem[{Gao et~al.(2023)Gao, Tow, Abbasi, Biderman, Black, DiPofi, Foster, Golding, Hsu, Le~Noac'h, Li, McDonell, Muennighoff, Ociepa, Phang, Reynolds, Schoelkopf, Skowron, Sutawika, Tang, Thite, Wang, Wang, and Zou}]{eval-harness}
Gao, L.; Tow, J.; Abbasi, B.; Biderman, S.; Black, S.; DiPofi, A.; Foster, C.; Golding, L.; Hsu, J.; Le~Noac'h, A.; Li, H.; McDonell, K.; Muennighoff, N.; Ociepa, C.; Phang, J.; Reynolds, L.; Schoelkopf, H.; Skowron, A.; Sutawika, L.; Tang, E.; Thite, A.; Wang, B.; Wang, K.; and Zou, A. 2023.
\newblock A framework for few-shot language model evaluation.

\bibitem[{G{\"u}nther et~al.(2023)G{\"u}nther, Milliken, Geuter, Mastrapas, Wang, and Xiao}]{gunther2023jina}
G{\"u}nther, M.; Milliken, L.; Geuter, J.; Mastrapas, G.; Wang, B.; and Xiao, H. 2023.
\newblock Jina embeddings: A novel set of high-performance sentence embedding models.
\newblock \emph{arXiv preprint arXiv:2307.11224}.

\bibitem[{Hager et~al.(2024)Hager, Jungmann, Holland, Bhagat, Hubrecht, Knauer, Vielhauer, Makowski, Braren, Kaissis et~al.}]{hager2024evaluation}
Hager, P.; Jungmann, F.; Holland, R.; Bhagat, K.; Hubrecht, I.; Knauer, M.; Vielhauer, J.; Makowski, M.; Braren, R.; Kaissis, G.; et~al. 2024.
\newblock Evaluation and mitigation of the limitations of large language models in clinical decision-making.
\newblock \emph{Nature medicine}, 1--10.

\bibitem[{He et~al.(2024)He, Huang, Jiang, Nie, Wang, Wang, and Chen}]{he2024foundation}
He, Y.; Huang, F.; Jiang, X.; Nie, Y.; Wang, M.; Wang, J.; and Chen, H. 2024.
\newblock Foundation model for advancing healthcare: Challenges, opportunities, and future directions.
\newblock \emph{arXiv preprint arXiv:2404.03264}.

\bibitem[{Jin et~al.(2021)Jin, Pan, Oufattole, Weng, Fang, and Szolovits}]{jin2021disease}
Jin, D.; Pan, E.; Oufattole, N.; Weng, W.-H.; Fang, H.; and Szolovits, P. 2021.
\newblock What disease does this patient have? a large-scale open domain question answering dataset from medical exams.
\newblock \emph{Applied Sciences}, 11(14): 6421.

\bibitem[{Jin et~al.(2019)Jin, Dhingra, Liu, Cohen, and Lu}]{jin2019pubmedqa}
Jin, Q.; Dhingra, B.; Liu, Z.; Cohen, W.~W.; and Lu, X. 2019.
\newblock Pubmedqa: A dataset for biomedical research question answering.
\newblock \emph{arXiv preprint arXiv:1909.06146}.

\bibitem[{Kanjee, Crowe, and Rodman(2023)}]{kanjee2023accuracy}
Kanjee, Z.; Crowe, B.; and Rodman, A. 2023.
\newblock Accuracy of a generative artificial intelligence model in a complex diagnostic challenge.
\newblock \emph{Jama}, 330(1): 78--80.

\bibitem[{Kim et~al.(2024)Kim, Hwang, Lee, Park, Kim, Lee, Yoon, Sohn, Choi, and Kang}]{kim2024small}
Kim, H.; Hwang, H.; Lee, J.; Park, S.; Kim, D.; Lee, T.; Yoon, C.; Sohn, J.; Choi, D.; and Kang, J. 2024.
\newblock Small language models learn enhanced reasoning skills from medical textbooks.
\newblock \emph{arXiv preprint arXiv:2404.00376}.

\bibitem[{Krithara et~al.(2023)Krithara, Nentidis, Bougiatiotis, and Paliouras}]{krithara2023bioasq}
Krithara, A.; Nentidis, A.; Bougiatiotis, K.; and Paliouras, G. 2023.
\newblock BioASQ-QA: A manually curated corpus for Biomedical Question Answering.
\newblock \emph{Scientific Data}, 10(1): 170.

\bibitem[{Labrak et~al.(2024)Labrak, Bazoge, Morin, Gourraud, Rouvier, and Dufour}]{labrak2024biomistral}
Labrak, Y.; Bazoge, A.; Morin, E.; Gourraud, P.-A.; Rouvier, M.; and Dufour, R. 2024.
\newblock Biomistral: A collection of open-source pretrained large language models for medical domains.
\newblock \emph{arXiv preprint arXiv:2402.10373}.

\bibitem[{Liu et~al.(2024)Liu, Zhou, Hua, Rohanian, Thakur, Clifton, and Clifton}]{liu2024large}
Liu, F.; Zhou, H.; Hua, Y.; Rohanian, O.; Thakur, A.; Clifton, L.; and Clifton, D.~A. 2024.
\newblock Large Language Models in the Clinic: A Comprehensive Benchmark.
\newblock \emph{medRxiv}, 2024--04.

\bibitem[{Liu et~al.(2023)Liu, Zhong, Li, Yang, Ju, Wu, Ma, Shu, Chen, Kim et~al.}]{liu2023tailoring}
Liu, Z.; Zhong, A.; Li, Y.; Yang, L.; Ju, C.; Wu, Z.; Ma, C.; Shu, P.; Chen, C.; Kim, S.; et~al. 2023.
\newblock Tailoring large language models to radiology: A preliminary approach to llm adaptation for a highly specialized domain.
\newblock In \emph{International Workshop on Machine Learning in Medical Imaging}, 464--473. Springer.

\bibitem[{Luo et~al.(2023)Luo, Zhang, Fan, Yang, Wu, Qiao, and Nie}]{luo2023biomedgpt}
Luo, Y.; Zhang, J.; Fan, S.; Yang, K.; Wu, Y.; Qiao, M.; and Nie, Z. 2023.
\newblock Biomedgpt: Open multimodal generative pre-trained transformer for biomedicine.
\newblock \emph{arXiv preprint arXiv:2308.09442}.

\bibitem[{Mart{\'\i}nez-deMiguel et~al.(2022)Mart{\'\i}nez-deMiguel, Segura-Bedmar, Chac{\'o}n-Solano, and Guerrero-Aspizua}]{martinez2022raredis}
Mart{\'\i}nez-deMiguel, C.; Segura-Bedmar, I.; Chac{\'o}n-Solano, E.; and Guerrero-Aspizua, S. 2022.
\newblock The RareDis corpus: a corpus annotated with rare diseases, their signs and symptoms.
\newblock \emph{Journal of Biomedical Informatics}, 125: 103961.

\bibitem[{Martviset et~al.(2020)Martviset, Sirisabhabhorn, Pumpa, Rhongbutsri, Taylor, and Taylor}]{martviset2020urinary}
Martviset, P.; Sirisabhabhorn, K.; Pumpa, S.; Rhongbutsri, P.; Taylor, A.; and Taylor, W.~R. 2020.
\newblock Urinary balantidiasis in a patient with systemic lupus erythematosus and lupus nephritis: a case report.
\newblock \emph{Journal of Medical Case Reports}, 14: 1--5.

\bibitem[{Mehandru et~al.(2024)Mehandru, Miao, Almaraz, Sushil, Butte, and Alaa}]{mehandru2024evaluating}
Mehandru, N.; Miao, B.~Y.; Almaraz, E.~R.; Sushil, M.; Butte, A.~J.; and Alaa, A. 2024.
\newblock Evaluating large language models as agents in the clinic.
\newblock \emph{NPJ digital medicine}, 7(1): 84.

\bibitem[{Nori et~al.(2023)Nori, King, McKinney, Carignan, and Horvitz}]{nori2023capabilities}
Nori, H.; King, N.; McKinney, S.~M.; Carignan, D.; and Horvitz, E. 2023.
\newblock Capabilities of gpt-4 on medical challenge problems.
\newblock \emph{arXiv preprint arXiv:2303.13375}.

\bibitem[{Ortu et~al.(2013)Ortu, Pietropaoli, Baldi, Marzo, Giannoni, and Monaco}]{ortu2013polyarteritis}
Ortu, E.; Pietropaoli, D.; Baldi, M.; Marzo, G.; Giannoni, M.; and Monaco, A. 2013.
\newblock Polyarteritis nodosa involving the hard palate: a case report.
\newblock \emph{Journal of medical case reports}, 7: 1--3.

\bibitem[{Pal, Umapathi, and Sankarasubbu(2022)}]{pal2022medmcqa}
Pal, A.; Umapathi, L.~K.; and Sankarasubbu, M. 2022.
\newblock Medmcqa: A large-scale multi-subject multi-choice dataset for medical domain question answering.
\newblock In \emph{Conference on health, inference, and learning}, 248--260. PMLR.

\bibitem[{Reese et~al.(2024)Reese, Chimirri, Danis, Caufield, Wissink, Casiraghi, Valentini, Haendel, Mungall, and Robinson}]{reese2024evaluation}
Reese, J.~T.; Chimirri, L.; Danis, D.; Caufield, J.~H.; Wissink, K.~W.; Casiraghi, E.; Valentini, G.; Haendel, M.~A.; Mungall, C.~J.; and Robinson, P.~N. 2024.
\newblock Evaluation of the Diagnostic Accuracy of GPT-4 in Five Thousand Rare Disease Cases.
\newblock \emph{medRxiv}, 2024--07.

\bibitem[{R{\'\i}os-Hoyo et~al.(2024)R{\'\i}os-Hoyo, Shan, Li, Pearson, Pusztai, and Howard}]{rios2024evaluation}
R{\'\i}os-Hoyo, A.; Shan, N.~L.; Li, A.; Pearson, A.~T.; Pusztai, L.; and Howard, F.~M. 2024.
\newblock Evaluation of large language models as a diagnostic aid for complex medical cases.
\newblock \emph{Frontiers in Medicine}, 11: 1380148.

\bibitem[{Shoham and Rappoport(2024{\natexlab{a}})}]{shoham2024cpllm}
Shoham, O.~B.; and Rappoport, N. 2024{\natexlab{a}}.
\newblock Cpllm: Clinical prediction with large language models.
\newblock \emph{PLOS Digital Health}, 3(12): e0000680.

\bibitem[{Shoham and Rappoport(2024{\natexlab{b}})}]{shoham2024medconceptsqa}
Shoham, O.~B.; and Rappoport, N. 2024{\natexlab{b}}.
\newblock MedConceptsQA--Open Source Medical Concepts QA Benchmark.
\newblock \emph{arXiv preprint arXiv:2405.07348}.

\bibitem[{Singhal et~al.(2023)Singhal, Azizi, Tu, Mahdavi, Wei, Chung, Scales, Tanwani, Cole-Lewis, Pfohl et~al.}]{singhal2023large}
Singhal, K.; Azizi, S.; Tu, T.; Mahdavi, S.~S.; Wei, J.; Chung, H.~W.; Scales, N.; Tanwani, A.; Cole-Lewis, H.; Pfohl, S.; et~al. 2023.
\newblock Large language models encode clinical knowledge.
\newblock \emph{Nature}, 620(7972): 172--180.

\bibitem[{Zhang et~al.(2019)Zhang, Kishore, Wu, Weinberger, and Artzi}]{zhang2019bertscore}
Zhang, T.; Kishore, V.; Wu, F.; Weinberger, K.~Q.; and Artzi, Y. 2019.
\newblock Bertscore: Evaluating text generation with bert.
\newblock \emph{arXiv preprint arXiv:1904.09675}.

\bibitem[{Zhou et~al.(2023)Zhou, Gu, Zou, Li, Chen, Zhou, Liu, Hua, Mao, Wu et~al.}]{zhou2023survey}
Zhou, H.; Gu, B.; Zou, X.; Li, Y.; Chen, S.~S.; Zhou, P.; Liu, J.; Hua, Y.; Mao, C.; Wu, X.; et~al. 2023.
\newblock A survey of large language models in medicine: Progress, application, and challenge.
\newblock \emph{arXiv preprint arXiv:2311.05112}.

\end{thebibliography}

\end{document}